\documentclass[sigconf]{acmart}

\usepackage{amsmath,tabularx}
\usepackage{amsthm}

\DeclareMathOperator*{\argmax}{arg\,max}

\usepackage[most]{tcolorbox}
\usepackage{multicol}
\usepackage{bbm}
\usepackage{enumitem}

\newtheorem{definition1}{Definition}

\usepackage{tikz}
\usetikzlibrary{decorations.pathreplacing,arrows,automata,calc,shapes.geometric,quotes}

\usepackage{bm}
\newcommand\ci{\perp\!\!\!\perp}

\acmConference[]{}{}{}

\begin{document}

\title[Counterfactual Formulation of Patient-Specific Root Causes of Disease]{Counterfactual Formulation of\\ Patient-Specific Root Causes of Disease}

\author{Eric V. Strobl}
\email{}
 \affiliation{%
   \institution{}
   \city{}
   \state{}
   \country{}
 }


\renewcommand{\shortauthors}{Eric V. Strobl}

\begin{abstract}
Root causes of disease intuitively correspond to root vertices that increase the likelihood of a diagnosis. This description of a root cause nevertheless lacks the rigorous mathematical formulation needed for the development of computer algorithms designed to automatically detect root causes from data. Prior work defined patient specific root causes of disease using an interventionalist account that only climbs to the second rung of Pearl's Ladder of Causation. In this theoretical piece, we climb to the third rung by proposing a counterfactual definition matching clinical intuition based on fixed factual data alone. We then show how to assign a root causal contribution score to each variable using Shapley values from explainable artificial intelligence. The proposed counterfactual formulation of patient-specific root causes of disease accounts for noisy labels, adapts to disease prevalence and admits fast computation without the need for counterfactual simulation. 
\end{abstract}


\maketitle

\section{Introduction}

Root causes of disease intuitively correspond to root vertices that increase the likelihood of a diagnosis. Clinicians assess for root causes by comparing the information they gather about a patient to their schemata, or preconceived notions of the world \cite{Piaget14}. Clinicians therefore implicitly ask themselves,
``Did learning that $X_i = x_i$ increase my belief that $x_i$ induced disease in this patient?'' where $x_i$ denotes a random insult such as a virus, mutation or traumatic event identified by clinical assessment, or \textit{backtracking} through the patient's history. Clinicians assume that $X_i$ is a root vertex, and the increase is relative to a ``typical person'' corresponding to their person schema just before they knew the value of $X_i$. If the answer to the question is affirmative, then clinicians conclude that $X_i$ is patient-specific root cause of disease. 

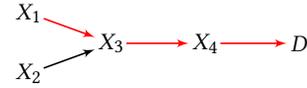
\begin{figure}[]
\centering
\begin{tikzpicture}[scale=1.0, shorten >=1pt,auto,node distance=2.8cm, semithick,
  inj/.pic = {\draw (0,0) -- ++ (0,2mm) 
                node[minimum size=2mm, fill=red!60,above] {}
                node[draw, semithick, minimum width=2mm, minimum height=5mm,above] (aux) {};
              \draw[thick] (aux.west) -- (aux.east); 
              \draw[thick,{Bar[width=2mm]}-{Hooks[width=4mm]}] (aux.center) -- ++ (0,4mm) coordinate (-inj);
              }]
                    
\tikzset{vertex/.style = {inner sep=0.4pt}}
\tikzset{edge/.style = {->,> = latex'}}
 
\node[vertex] (2) at  (0.15,-0.4) {$X_2$};
\node[vertex] (3) at  (1.25,0) {$X_3$};
\node[vertex] (4) at  (2.5,0) {$X_4$};
\node[vertex] (5) at  (3.75,0) {$D$};
\node[vertex] (1) at  (0.15,0.4) {$X_1$};

\draw[edge,red] (1) to (3);
\draw[edge] (2) to (3);
\draw[edge,red] (3) to (4);
\draw[edge,red] (4) to (5);
\end{tikzpicture}
\setlength{\belowcaptionskip}{-5pt}
\caption{A root causal inference process employed by clinicians. The sudden death of a spouse $x_1$ leads to depression $x_3$, alcohol use $x_4$ and then cirrhosis $D=1$. $X_1$ is the root cause, and $X_2$ denotes inconsequential family history of depression.} \label{fig_shock}
\end{figure}

The following is a simplified but representative example focusing on one root cause, even though a patient may have multiple root causes of disease in practice. A patient visits a physician after noticing jaundice, or yellowing of the skin. The patient suddenly lost his wife to a car accident, became depressed, started drinking alcohol and then developed cirrhosis (scarring of the liver). He also tells the physician that his grandmother had depression when he was a child. We can represent this causal process as the directed graph shown in Figure \ref{fig_shock}, where $D$ denotes the diagnosis of cirrhosis, $\bm{X}$ the set of upstream variables, and directed edges the direct causal relations. The variable $X_1$ represents the status of the spouse, $X_2$ the family history of depression, $X_3$ the patient's depression severity, and $X_4$ the amount of alcohol use. 

The physician knows that the sudden unexpected loss of a spouse $x_1$ frequently leads to depression, which in turn can lead to excessive alcohol use and then cirrhosis. The physician concludes that $X_1$ is the root cause of the patient's cirrhosis because (1) $X_1$ is a root vertex and (2) knowing $X_1 
= x_1$ substantially increases the likelihood of the patient developing disease relative to an imagined typical person. In contrast, knowing $x_2$ adds little value because few individuals with a remote family history of depression develop depression themselves. Notice that the physician infers the root cause by backtracking on inferences made from \textit{fixed factual data} because his concept of a ``typical person'' comes from his and others' factual lived experiences. The physician does \textit{not} infer the root cause by asking the patient or himself questions that explicitly construct interventions in counterfactual worlds, as commonly suggested in the causal inference literature \cite{Pearl09}.\footnote{Had the physician intervened to prevent the wife's death counter to the fact, then the patient would probably not have developed cirrhosis -- not because of the imagined intervention, but because learning of the wife's presence does not increase the physician's belief that the wife's presence induced disease in this patient.}

The physician can avoid interventions because if $P(D|x_i) \not = \mathbb{E}(P(D|X_i))$ for any root vertex $X_i$, where the left hand side denotes the patient with $x_i$ and the right hand side the typical person, then $X_i$ causes $D$ by the global Markov property (see Section \ref{sec:causal_models}). Some authors exploited this fact to identify patient-specific root causes of disease and quantify their effects using only the \textit{factual} (or non-interventional) values of the error terms $\bm{E}$ of structural equation models \cite{Strobl22a,Strobl22b,Strobl23}. Each error term $E_i \in \bm{E}$ is a root vertex representing the stochasticity of the observed variable $X_i \in \bm{X}$ given its direct causes. The error term values correspond to exogenous but possibly observed insults, such as the spouse's death $E_1 = X_1 = x_1$, that can induce changes in the other variables, such as depression severity $X_3$. We can also interpret the error terms as natural stochastic interventions; they are \textit{natural} and \textit{stochastic} because they are drawn from $\mathbb{P}(\bm{E})$ with mutual independence by nature, and they are \textit{interventions} because the error terms have no direct causes.\footnote{Some authors assume that only humans can perform interventions. We speak more generally: \textit{nature} can intervene on the observed and latent error terms, but \textit{humans} can only intervene on (some subset of) the observed error terms.} Investigators proposed to quantify the causal contribution of the error term value $E_i = e_i$ on $D$ using a Shapley value based on conditional expectations of models that predict $D$ \cite{Lundberg17}. If $e_i$ is associated with a positive Shapley value, then $X_i$ is a patient-specific root cause of disease.

This formulation of patient-specific root causes of disease as predictive, natural and stochastic interventions unfortunately fails to align with the \textit{counterfactual} nature of clinical reasoning, where clinicians compare each patient against an imagined typical person. Pearl's Ladder of Causation constitutes a hierarchy of three problems of increasing difficulty: (1) prediction, (2) intervention and (3) counterfactuals \cite{Pearl09}. Prediction identifies the probability of $D$ conditional on any $\bm{V} \subseteq \bm{X}$. Intervention identifies the probability of $D$ after manually modifying $\bm{V}$ given $\bm{W} \subseteq \textcolor{red}{\bm{X} \setminus \bm{V}}$. Counterfactuals identify the probability of $D$ after manually modifying $\bm{V}$ given $\bm{W} \subseteq \textcolor{red}{\bm{X}}$. Solving a rung of the ladder thus solves all lower rungs, but the reverse fails. The existing definition of a patient-specific root cause of disease only applies to the second rung. We instead seek a counterfactual formulation of patient-specific root causes.

In this paper, we introduce a definition of patient-specific root causes of disease based on counterfactuals that corresponds with clinical intuition on fixed factual data. We leverage recent work in the formalization of backtracking counterfactuals that allows us to equate interventions to the factual values of the error terms in structural equation models \cite{Von23}. We also demonstrate how to design automated procedures that quantify the root causal contribution of each error term value using the Shapley values of \cite{Lundberg17}. The definition strengthens the justification of a number of existing algorithms utilized for patient-specific root causal inference \cite{Strobl22a,Strobl22b,Strobl23}.

\section{Background}
\subsection{Causal Models} \label{sec:causal_models}
A \textit{causal model} corresponds to the triple $\mathcal{M} = \langle \bm{E}, \bm{X}, \mathbb{F} \rangle$. The set $\bm{E}$ contains $p$ \textit{exogenous error terms} determined by factors outside the model. The set $\bm{X}$ contains $p$ \textit{endogenous variables}. Each variable in $\bm{U} = \bm{X} \cup \bm{E}$ is determined by the functions $\mathbb{F}$ and error terms. In particular, each function $f_{U_i} \in \mathbb{F}$ takes $E_i \cup \textnormal{Pa}(U_i)$ as input, where $\textnormal{Pa}(U_i) \subseteq \bm{X} \setminus U_i$ is the \textit{parent set} of $U_i$, and deterministically outputs $U_i$ via the \textit{structural equation model} (SEM):
\begin{equation}
\begin{aligned}
    U_i &= f_{U_i}(\textnormal{Pa}(U_i), E_i) \hspace{5mm} \forall U_i \in \bm{U}.
\end{aligned}
\end{equation}
A \textit{root vertex} is a vertex with no parents. All error terms are root vertices, and we assume that $E_i = f_{E_i}(E_i)$ and $X_j =  f_{X_j}(E_j)=E_j$ when $X_j$ is a root vertex without loss of generality. A \textit{sink vertex} is a vertex that is not a member of any parent set. The set of functions $\mathbb{F}$ outputs a unique set of values of $\bm{X}$ given $\bm{E} = \bm{e}$, denoted by $\bm{X}(\bm{e})$. An SEM is \textit{invertible} if we can recover $\bm{e}$ uniquely from $\bm{X}(\bm{e})$. 

We can associate a \textit{directed graph} $\mathbb{G}(\mathcal{M})$ to the causal model $\mathcal{M}$ by drawing a directed edge from $\textnormal{Pa}(X_i) \cup E_i$ to $X_i$ for each $X_i \in \bm{X}$. Investigators sometimes do not include $\bm{E}$ in the directed graph. A \textit{directed path} from $U_i$ to $U_j$ corresponds to a sequence of adjacent directed edges from $U_i$ to $U_j$. A subset $\bm{W} \subseteq \bm{U}$ is an ancestor of $U_i$ if there exists a directed path from at least one member of $\bm{W}$ to $U_i$ (or $U_i \in \bm{W}$). A \textit{collider} refers to $U_j$ in the triple $U_i \rightarrow U_j \leftarrow U_k$. Two vertices $U_i$ and $U_j$ are \textit{d-connected} given $\bm{W} \setminus \{U_i,U_j\}$ if there exists a path between $U_i$ and $U_j$ such that every collider is an ancestor of $\bm{W}$ and no non-collider is in $\bm{W}$. The two vertices are likewise \textit{d-separated} if they are not d-connected. The directed graph $\mathbb{G}(\mathcal{M})$ is \textit{acyclic} when a directed path exists from $U_i$ to $U_j$, but we do not have the directed edge $U_j \rightarrow U_i$ for any two distinct vertices $U_i, U_j \in \bm{U}$. We only consider acyclic graphs in this paper.

A \textit{causal submodel} $\mathcal{M}_{\bm{V}}$ of $\mathcal{M}$ corresponds to the triple $\mathcal{M}_{\bm{V}} = \langle \bm{E}, \bm{X}, \mathbb{F}_{\bm{V}} \rangle$ where $\mathbb{F}_{\bm{V}} = \{ f_{U_i} : U_i \not \in \bm{V} \} \cup \bm{V}$ and $\bm{V} \subseteq \bm{U}$. The effect of the action $\textnormal{do}(\bm{V}=\bm{v})$ corresponds to the submodel $\mathcal{M}_{\bm{V} = \bm{v}}$, or more succinctly $\mathcal{M}_{\bm{v}}$, with the constants $\bm{v}$. However, we may more generally insert possibly non-constant random variables $\bm{V}$ with the action $\textnormal{do}(\bm{V})$ corresponding to the submodel $\mathcal{M}_{\bm{V}}$. A \textit{causal world} refers to the pair $\langle \mathcal{M}, \bm{e} \rangle$, where $\bm{e}$ is a particular realization of the error terms $\bm{E}$. The \textit{potential outcome} of $\bm{Y} \subseteq \bm{X}$ to the action $\textnormal{do}(\bm{V})$ in the world $\langle \mathcal{M}, \bm{e} \rangle$ is denoted by $\bm{Y}_{\bm{V}}(\bm{e})$, or the output of $\mathbb{F}_{\bm{V}}$ for $\bm{Y}$ given $\bm{e}$. The \textit{counterfactual sentence}, ``$\bm{Y}$ would be $\bm{y}$ in situation $\bm{e}$ had we introduced $\bm{V}$'' corresponds to  $\bm{Y}_{\bm{V}}(\bm{e}) = \bm{y}.$

A \textit{probabilistic causal model} refers to the pair $\langle \mathcal{M}, \mathbb{P}(\bm{E}) \rangle$. We focus on Markovian models where the probability distribution factorizes into $\prod_{i=1}^p \mathbb{P}(E_i)$ so that $\bm{E}$ more specifically corresponds to a set of  mutually independent error terms. Markovian models satisfy the \textit{global Markov property}, where d-separation between $U_i$ and $U_j$ given $\bm{W} \setminus \{U_i,U_j\}$ implies that $U_i$ and $U_j$ are conditionally independent given $\bm{W}$ \cite{Lauritzen90}; we denote the conditional independence by $U_i \ci U_j | \bm{W}$. The term \textit{d-separation faithfulness} corresponds to the converse, where conditional independence implies d-separation. The joint distribution $\mathbb{P}(\bm{E})$ together with the functions $\mathbb{F}_{\bm{V}}$ induces a distribution over any subset of the endogenous variables:
$\mathbb{P}(\bm{Y}_{\bm{V}} = \bm{y}) = \sum_{\bm{e}} \mathbb{P}(\bm{e}) \bm{1}_{\bm{Y}_{\bm{V}}(\bm{e}) = \bm{y}}.$
We use the summation symbol to represent either summation over a probably mass function in the discrete case or integration over a probably density function in the continuous case in order to ease notation and avoid measure theoretic details. 

\subsection{Interventional Counterfactuals}

We now notationally distinguish the factual values $\bm{e}$ from the counterfactual values $\bm{e}^*$ and likewise for random variables. We pay special attention to computing counterfactual distributions of the form $\bm{P}(\bm{Y}^*_{\bm{V}^*}=\bm{y}^* | \bm{z})$ given the factual values of $\bm{Z} \subseteq \bm{U}$. The interventional approach to computing counterfactuals involves three steps:
\begin{enumerate}[leftmargin=*,label=(\arabic*)]
    \item Abduction: update $\mathbb{P}(\bm{E})$ by the evidence $\bm{z}$ to obtain $\mathbb{P}(\bm{E}|\bm{z})$.
    \item Action: modify $\mathcal{M}$ with the action $\bm{V}^*=\textnormal{do}(\bm{V})$ to obtain the submodel $\mathcal{M}_{\bm{V}^*}$.
    \item Prediction: use the probabilistic causal model $\langle \mathcal{M}_{\bm{V}^*}, \mathbb{P}(\bm{E}|\bm{z}) \rangle$ to obtain $\mathbb{P}(\bm{Y}^*_{\bm{V}^*} | \bm{z})$.
\end{enumerate}

The interventional approach therefore assumes that the error term distribution $\mathbb{P}(\bm{E}|\bm{z})$ is shared between the \textit{factual causal model} $\langle \mathcal{M}, \mathbb{P}(\bm{E}) \rangle$ and the \textit{counterfactual causal model} $\langle \mathcal{M}_{\bm{V}^*}, \mathbb{P}(\bm{E}|\bm{z}) \rangle$. Interventions modify the functions $\mathbb{F}$ to $\mathbb{F}_{\bm{V}^*}$.

\subsection{Backtracking Counterfactuals} \label{sec:backtrack}

Backtracking counterfactuals take an alternative approach by assuming that the functions remain intact between the factual and counterfactual models, but the distribution $\mathbb{P}(\bm{E}|\bm{z})$ can differ. In the present context, the term ``backtracking'' refers to the process of updating the error term distributions in order to explain counterfactual distributions. 

Multiple different error term distributions may explain a counterfactual distribution, so backtracking is not unique. The non uniqueness prompted \cite{Von23} to introduce the backtracking conditional, or the distribution $\mathbb{P}(\bm{E}^* | \bm{E})$. The backtracking conditional given $\bm{e}$, or $\mathbb{P}(\bm{E}^* | \bm{E}=\bm{e})$, refers to the likelihood of counterfactual values $\bm{e}^*$ given the factual values $\bm{e}$. As a result, the backtracking conditional offers a flexible framework for encoding notions of cross-model similarity between the error term distributions.

We compute backtracking counterfactuals using the following three steps akin to the three steps of interventional counterfactuals:
\begin{enumerate}[leftmargin=*,label=(\arabic*)]
    \item Abduction: update $\mathbb{P}(\bm{E}^*,\bm{E})$ by the evidence $(\bm{v}^*,\bm{z})$ to obtain $\mathbb{P}(\bm{E}^*,\bm{E} | \bm{v}^*, \bm{z}) = \frac{\mathbb{P}(\bm{E}^*,\bm{E})}{\mathbb{P}(\bm{v}^*, \bm{z})} \bm{1}_{\bm{V}^*(\bm{E}^*)=\bm{v}^*} \bm{1}_{\bm{Z}(\bm{E})=\bm{z}}$.
    \item Marginalization: marginalize out $\bm{E}$ to obtain $\mathbb{P}(\bm{E}^* | \bm{v}^*, \bm{z}) =$\\ $\sum_{\bm{e}} \mathbb{P}(\bm{E}^*,\bm{e} | \bm{v}^*, \bm{z}).$
    \item Prediction: use the probabilistic causal model  $\langle \mathcal{M}, \mathbb{P}(\bm{E}^* | \bm{v}^*, \bm{z}) \rangle$ to obtain $\mathbb{P}(\bm{Y}^*|\bm{v}^*,\bm{z}) = \sum_{\bm{e}^*} \mathbb{P}(\bm{e}^*|\bm{v}^*,\bm{z}) \bm{1}_{\bm{Y}^*(\bm{e}^*)}.$
\end{enumerate}
We therefore arrive at the backtracking counterfactual distribution $\mathbb{P}(\bm{Y}^*|\bm{v}^*,\bm{z})$ similar to the interventional counterfactual distribution $\mathbb{P}(\bm{Y}^*_{\bm{V}^*} | \bm{z})$.

The choice of the backtracking conditional $\mathbb{P}(\bm{E}^* | \bm{E})$ depends on the area of application. The authors in \cite{Von23} do not specify the form of the conditional distribution in the context of biomedical applications, but they provide three desiderata in the general case. The backtracking conditional should satisfy all of the following for any $\bm{e}$ and $\bm{e}^*$: (1) closeness: $\bm{e} = \argmax_{\bm{e}^*} \mathbb{P}(\bm{e}^* | \bm{e})$, (2) symmetry: $\mathbb{P}(\bm{e}^* | \bm{e}) = \mathbb{P}(\bm{e} | \bm{e}^*)$, (3) decomposability: $\mathbb{P}(\bm{e}^* | \bm{e}) = \prod_{i=1}^p \mathbb{P}(e_i^* | e_i)$. We will choose a backtracking conditional that satisfies these three properties in Section \ref{sec:root_back}. 

\section{Root Causes}

\subsection{Approach}
We consider an invertible SEM over $\bm{X}$ and introduce an additional endogenous variable $D$ indicating the diagnosis. The diagnosis $D$ is binary, where we have $D=1$ for a patient deemed to have disease and $D=0$ for a healthy control. The diagnosis is a noisy label in general, since it may differ slightly between diagnosticians in practice. 

We further assume that $D$ is a sink vertex so that $D$ is not a parent of any vertex in $\bm{X}$. This assumption is reasonable because $\bm{X}$ often contains variables representing entities like images, gene expression levels, environmental factors or laboratories. Investigators thus believe that these variables are instantiated before the diagnosis in time.

We seek to identify the patient-specific root causes of $D$ in $\bm{X}$; it is not informative to claim that $D$ is a patient-specific root cause of the same endogenous variable $D$. We therefore reserve the notation $\bm{X}$ for the other endogenous variables denoting patient characteristics and $\bm{E}$ for the error terms of $\bm{X}$ so that $D \not \in \bm{X}$ and $E_D \not \in \bm{E}$. We refer to a patient by the instantiation $\bm{X}(\bm{e}) = \bm{x}$, or equivalently $\bm{e}$ with SEMs invertible over $\bm{X}$. We henceforth only implicitly assume the presence of $E_D$ to prevent cluttering of notation. The model $\langle \mathcal{M}, \mathbb{P}(\bm{E}) \rangle$ thus more specifically means $\langle \mathcal{M}, \mathbb{P}(E_D, \bm{E}) \rangle$, and the likewise the world $\langle \mathcal{M}, \bm{e} \rangle$ means $\langle \mathcal{M}, e_D \cup \bm{e} \rangle$.

In this section, we will derive an interventional and similar backtracking counterfactual formulation of patient-specific root causes of disease \textcolor{red}{for a set of values $\bm{e}_{\bm{v}} \subseteq \bm{e}$} using three steps:
\begin{enumerate}[leftmargin=*,label=(\arabic*)]
    \item Define root causes of $D$ as root vertices with appreciable causal effects on $D$;
    \item Use interventional counterfactuals to quantify root causal effects with factual and counterfactual worlds that preserve error term \textit{distributions};
    \item Equate the interventional counterfactual formulation to a backtracking one that preserves error term \textit{values} between the worlds -- thus matching clinical intuition.
\end{enumerate}
We will then derive a measure of patient-specific root causal contribution for each \textit{single value} $e_i \in \bm{e}$ regardless of the choice of the set $\bm{e}_{\bm{v}}$ containing $e_i$ in Section \ref{sec:contrib}, but the derivation will depend on the arguments presented in this section.

\subsection{From Root Vertices}

We say that $\bm{V} \subseteq \bm{U}$ is an \textit{appreciable cause} of $D$ if $\bm{V}$ only contains ancestors of $D$ and $\bm{V} \not \ci D$. The conditional dependence relation is implied by d-separation faithfulness, which we do not assume. Likewise, consider the quantity:
\begin{equation} \nonumber
    \Phi_{\bm{V}} = \mathbb{P}(D|\bm{V}) - \mathbb{P}(D) = \mathbb{P}(D|\bm{V}) - \mathbb{E}_{\bm{V}}(\mathbb{P}(D|\bm{V})).
\end{equation}
So that $\bm{V}$ is an appreciable cause of $D$ \textcolor{red}{for $\bm{V}=\bm{v}$} if $\bm{V}$ only contains ancestors of $D$ and $\Phi_{\bm{V}=\bm{v}} \not = 0$. Moreover, $\Phi_{\bm{v}}$ quantifies the discrepancy between $\mathbb{P}(D|\bm{v})$ and $\mathbb{P}(D)$ and therefore corresponds to a measure of the causal effect of $\bm{v}$ on $D$.

Assume further that $\bm{V}$ only contains root vertices so that $\bm{V}=\bm{E}_{\bm{V}}$, or the set of error terms associated with $\bm{V}$. We again quantify the causal effect of $\bm{V}$ by:
\begin{equation}
    \Phi_{\bm{V}} = \mathbb{P}(D|\bm{E}_{\bm{V}}) - \mathbb{E}_{\textcolor{red}{\bm{E}_{\bm{V}}}}(\mathbb{P}(D|\bm{E}_{\bm{V}})),
\end{equation}
where we emphasize that we average over $\bm{E}_{\bm{V}}$. Thus the set of root vertices is an appreciable cause of $D$ for $\bm{v}$ if $\Phi_{\bm{v}} \not = 0$; we automatically have $\Phi_{\bm{v}} = 0$ when $\bm{E}_{\bm{V}}$ is not an ancestor of $D$ by the global Markov property. We make the following definition that only applies to root vertices:
\begin{definition1} \label{def:val}
    (Root cause) The set $\bm{V} \subseteq \bm{U}$ is a root cause of $D$ for $\bm{v}$ if $\bm{V}$ only contains root vertices and is an appreciable cause of $D$ for $\bm{v}$, or $\Phi_{\bm{v}} \not = 0$.
\end{definition1}

\subsection{As Counterfactual Interventions} \label{sec:interv_counter}

Definition \ref{def:val} only achieves specificity to $\bm{V} = \bm{v}$, but we seek specificity to an entire patient $\bm{x}$. Interventional counterfactuals provide an ideal framework for thinking about root causes as patient-specific, natural and stochastic interventions because we can explicitly enforce the do-operator using the counterfactual world\\ $\langle \mathcal{M}_{\bm{E}^*_{\bm{V}}}, \bm{e} \rangle$ for the patient $\bm{e}$; we introduce the \textit{stochastic} variable $\bm{E}^*_{\bm{V}} = \textnormal{do}(\bm{E}_{\bm{V}})$ with distribution $\mathbb{P}(\bm{E}^*_{\bm{V}}) = \mathbb{P}(\bm{E}_{\bm{V}})$. 

We proceed with the three steps of counterfactual interventions. The conditional distribution $\mathbb{P}(\bm{E}|\bm{x})$ in abduction has a point mass on $\bm{e}$ in invertible SEMs over $\bm{X}$. The counterfactual model\\ $\langle \mathcal{M}_{\bm{E}^*_{\bm{V}}}, \mathbb{P}(\bm{E}|\bm{x}) \rangle$ in the action step is thus equivalent to the \textit{counterfactual world} $\langle \mathcal{M}_{\bm{E}^*_{\bm{V}}}, \bm{e} \rangle$. Hence, we associate each patient $\bm{e}$ with the counterfactual world $\langle \mathcal{M}_{\bm{E}^*_{\bm{V}}}, \bm{e} \rangle$. 

We rewrite the counterfactual question in the Introduction as a double negative:
``Did not knowing the value of $X_i$ decrease my belief that $x_i$ induced disease in this patient?'' We then seek to answer the following analogous question in the prediction step: ``Would performing $\textnormal{do}(\bm{E}_{\bm{V}})$ have decreased the likelihood that patient $\bm{e}$ develops disease on average?'' The two questions are equivalent when $\bm{E}_{\bm{V}} = X_i$ because the do-operator corresponds to the physician's schema before knowing the values of $\bm{E}_{\bm{V}}$. In particular, we quantify the likelihood of the patient $\bm{e}$ developing disease in the factual world $\langle \mathcal{M}, \bm{e} \rangle$ via the conditional distribution $\mathbb{P}(D|\bm{e})$. We then answer the counterfactual question by comparing $\mathbb{P}(D|\bm{e})$ to the likelihood that the patient $\bm{e}$ develops disease in the counterfactual world where we do not know the values of $\bm{E}_{\bm{V}}$, i.e., by comparing the patient to a typical person corresponding to the average over $\bm{E}^*_{\bm{V}}$ in $\langle \mathcal{M}_{\bm{E}^*_{\bm{V}}}, \bm{e} \rangle$:
\begin{equation} \label{eq:interv_prob}
\begin{aligned}
\Phi_{\bm{e}_{\bm{V}}}(\bm{e})
&= \mathbb{P}(D|\bm{e}) - \mathbb{E}_{\bm{E}^*_{\bm{V}}}\mathbb{P}(D^*_{\bm{E}^*_{\bm{V}}}|\bm{e}_{\bm{W}}, \bm{E}^*_{\bm{V}})\\
&= \mathbb{P}(D|\bm{e}) - \mathbb{E}_{\bm{E}_{\bm{V}}}\mathbb{P}(D|\bm{e}_{\bm{W}}, \bm{E}_{\bm{V}}), 
\end{aligned}
\end{equation}
where $\bm{W} = \bm{X} \setminus \bm{V}$, and the last equality follows due to the equivalence of the error term distributions $\mathbb{P}(\bm{E}^*_{\bm{V}}) = \mathbb{P}(\bm{E}_{\bm{V}})$ between the worlds. We are now ready for a new definition:
\begin{definition1} \label{def:pt_root}
(Patient-specific root cause) The set $\bm{E}_{\bm{V}}$ is a patient-specific root cause of $D$ for $\bm{e}_{\bm{V}}$ if $\Phi_{\bm{e}_{\bm{V}}}(\bm{e}) \not =0$. Likewise $\bm{V} \subseteq \bm{X}$ is a patient-specific root cause of $D$ for $\bm{e}_{\bm{V}}$ projected onto $\bm{X}$. We more specifically say that $\bm{E}_{\bm{V}}$ is a patient-specific root cause \textit{of disease} ($D=1$) for $\bm{e}_{\bm{V}}$ if $\Phi_{\bm{e}_{\bm{V}}}(\bm{e})>0$, and likewise for the projection $\bm{V}$.
\end{definition1}
\noindent We encourage the reader to compare this definition against Definition \ref{def:val}. The second part of the above definition holds because $\Phi_{\bm{e}_{\bm{V}}}(\bm{e})>0$ implies that $\bm{e}_{\bm{V}}$ increases the probability of the patient developing disease relative to the counterfactual average. We consider projections onto observed variables because we may not observe all error terms.

\subsection{Extension to Backtracking} \label{sec:root_back}

Interventional counterfactuals require interventions on the error terms, so the counterfactual error term values can differ from their factual counterparts. This approach therefore fails to match clinical intuition that uses fixed factual data alone. We fix this wrinkle by equating interventional counterfactuals to backtracking counterfactuals that preserve error term values between worlds. 

We proceed with the three steps of backtracking counterfactuals. Abduction requires a choice for the backtracking conditional, so we set it to $\mathbb{P}(\bm{E}^* = \bm{e} | \bm{e}) = \prod_{i=1}^p \mathbb{P}(E_i^* = e_i | e_i) = 1$ in order to preserve error term values between worlds; this conditional satisfies the closeness, symmetry and decomposability desiderata. We proceed with abduction and marginalization given the evidence $(\bm{e}^*, \bm{e})$ yielding the conditional $\mathbb{P}(\bm{E}^* = \bm{e} | \bm{e}^*, \bm{e}) = \mathbb{P}(\bm{E}^* = \bm{e} | \bm{e}) = 1$. 

We now ask the same counterfactual question as in Section \ref{sec:interv_counter}, ``Would performing $\textnormal{do}(\bm{E}_{\bm{V}})$ have decreased the likelihood that patient $\bm{e}$ develops disease on average?'' This question is difficult to directly answer with backtracking counterfactuals, since we must perform an intervention. However, the choice of the backtracking conditional $\mathbb{P}(\bm{E}^* = \bm{e} | \bm{e}) = 1$ ensures the following equivalency:
\begin{equation} \label{eq:back1}
\begin{aligned}
 \mathbb{P}(D|\bm{e}) &= \sum_{\bm{e}^*} \mathbb{P}(\bm{E}^* =\bm{e}^*|\bm{e}) \bm{1}_{D^*(\bm{e}^*)=1},
\end{aligned}
\end{equation}
because the error term values remain unchanged between the factual and counterfactual worlds. We also have the following equivalence relation with interventional counterfactuals:
\begin{equation} \label{eq:back2}
\begin{aligned}
&\hspace{3.6mm} \mathbb{E}_{\bm{E}^*_{\bm{V}}}\mathbb{P}(D^*_{\bm{E}^*_{\bm{V}}}|\bm{e}_{\bm{W}}, \bm{E}^*_{\bm{V}}) = \mathbb{E}_{\bm{E}_{\bm{V}}}\mathbb{P}(D|\bm{e}_{\bm{W}}, \bm{E}_{\bm{V}}) \\ &= \mathbb{E}_{\bm{E}_{\bm{V}}}\sum_{\bm{e}^*} \mathbb{P}_{\bm{E}^*|\bm{E}}(\bm{e}^*|\bm{e}_{\bm{W}},\bm{E}_{\bm{V}}) \bm{1}_{D^*(\bm{e}^*)=1},
\end{aligned}
\end{equation}
where $\mathbb{P}_{\bm{E}^*|\bm{E}}(\bm{e}|\bm{e}) = 1$. Equations \eqref{eq:back1} and \eqref{eq:back2} thus ensure that we can identify $\Phi_{\bm{e}_{\bm{V}}}(\bm{e})$ in Equation \eqref{eq:interv_prob} and patient-specific root causes per Definition \ref{def:pt_root} \textit{without changing the error term values}. The backtracking interpretation is powerful because we do not need to explicitly enforce the do-operator, and it matches the way clinicians identify patient-specific root causes of disease by backtracking directly on the \textit{fixed factual data}.

\section{Root Causal Contributions} \label{sec:contrib}

Definition \ref{def:pt_root} suggests an algorithmic strategy for patient-specific root causal inference. We first identify the factual error term values from an invertible SEM over $\bm{X}$. We can recover the values either from the top down (root to sink vertices) in the linear case \cite{Strobl22a,Shimizu11}, or from the bottom up (sink to root vertices) in the general case \cite{Strobl22b,Peters14}. We then build a predictive model that recovers $\mathbb{P}(D|\bm{E})$ from data, so we can compute  $\Phi_{\bm{e}_{\bm{V}}}(\bm{e})$ for any choice of $\bm{V} \subseteq \bm{X}$ and $\bm{E} = \bm{e}$.

The quantity $\Phi_{\bm{e}_{\bm{V}}}(\bm{e})$ however only quantifies the root causal effect of the set $\bm{e}_{\bm{V}}$. Each clinician may imagine a different typical person depending on the choice of $\bm{E}_{\bm{V}}$. However, clinicians and patients ultimately want to communicate on a common ground, so they instead want to know the root causal contribution of each error term value \textit{regardless} of the choice of $\bm{E}_{\bm{V}}$. 

We now quantify the root causal contribution of each individual $e_i \in \bm{e}$ for a patient $\bm{e}$ regardless of the choice of $\bm{E}_{\bm{V}}$. We do so by first comparing the root causal effects of a set $\bm{e}_{\bm{W} \cup X_i}$ that includes $e_i$ and the corresponding set $\bm{e}_{\bm{W}}$ that does not:
\begin{equation} \nonumber
\begin{aligned}
   \gamma_{ \bm{e}_{\bm{W} \cup X_i}} &\triangleq \Phi_{\bm{e}_{\bm{V}}}(\bm{e}) - \Phi_{\bm{e}_{\bm{V} \setminus X_i}}(\bm{e})\\
   &= \mathbb{E}_{\bm{E}_{\bm{V} \setminus X_i}}\mathbb{P}(D|\textcolor{red}{\bm{e}_{\bm{W} \cup X_i}}, \bm{E}_{\bm{V}\setminus X_i}) - \mathbb{E}_{\bm{E}_{\bm{V}}}\mathbb{P}(D|\textcolor{red}{\bm{e}_{\bm{W}}}, \bm{E}_{\bm{V}}),
\end{aligned}
\end{equation}
where $\bm{E}_{\bm{W}} = \bm{E} \setminus \bm{E}_{\bm{V}}$. We do not prefer any particular set $\bm{e}_{\bm{W}} \subseteq (\bm{e} \setminus e_i)$ a priori. We therefore average over all possible combinations of $\bm{e}_{\bm{W}}$:
\begin{equation} \nonumber
 s_i = \frac{1}{p}\sum_{\bm{e}_{\bm{W}} \subseteq (\bm{e} \setminus e_i)} \frac{1}{\binom{p-1} {|\bm{e}_{\bm{W}}|}} \gamma_{\bm{e}_{\bm{W}\cup X_i}}.
\end{equation}
In other words, we compare all person schemas that include $e_i$ to those that do not, and then average them. 

The quantity $s_i$ is precisely the \textit{Shapley value} of \cite{Lundberg17} which satisfies three desiderata:
\begin{enumerate}[leftmargin=*,label=(\arabic*)]
    \item Local accuracy: $\sum_{i=1}^p s_i = \mathbb{P}(D|\bm{e}) - \mathbb{P}(D)$;
    \item Missingness: if $E_i \not \in \bm{E}$, then $s_i = 0$.
    \item Consistency: $s_i^\prime \geq s_i$ for any two distributions $\mathbb{P}^\prime$ and $\mathbb{P}$ where $\gamma^\prime_{\bm{e}_{\bm{W}\cup X_i}} \geq \gamma_{\bm{e}_{\bm{W}\cup X_i}}$ for all $\bm{E}_{\bm{W}} \subseteq (\bm{E} \setminus E_i)$.
\end{enumerate}
The first criterion ensures that the total score $\mathbb{P}(D|\bm{e}) - \mathbb{P}(D)$ distributed among the Shapley values remains invariant to changes in the disease prevalence rate $\mathbb{P}(D)$. Thus the total score of a patient $\bm{e}$ inferred from a population with high prevalence remains the same even when inferred from a population with low prevalence. The second criterion ensures that $s_D = 0$ for $E_D \not \in \bm{E}$. The third criterion means that, if a variable does not decrease the likelihood of disease in one population relative to another (for all sets), then its Shapley value does not decrease. The first and third criteria together imply that each Shapley value $s_i$ is also invariant to changes in the prevalence rate, where we have $\gamma^\prime_{\bm{e}_{\bm{W}\cup X_i}} = \gamma_{\bm{e}_{\bm{W}\cup X_i}}$ for all $\bm{E}_{\bm{W}} \subseteq (\bm{E} \setminus E_i)$ and $X_i \in \bm{X}$. These desiderata are thus necessary for any root causal contribution measure. The Shapley values are in fact the \textit{only} values satisfying the above three desiderata within the class of additive feature attribution measures \cite{Lundberg17}.

Note that we have used probability distributions in Equation \eqref{eq:interv_prob}, but we can likewise consider any strictly monotonic function $m$ of $\mathbb{P}$ such as the logarithm or logarithmic odds ratio to achieve the same idea: $\Phi^m_{\bm{e}_{\bm{V}}}(\bm{e}) = m\left[\mathbb{P}(D|\bm{e})\right] - \mathbb{E}_{\bm{E}_{\bm{V}}}m\left[\mathbb{P}(D|\bm{e}_{\bm{W}}, \bm{E}_{\bm{V}})\right].$ We then compute the corresponding Shapley values $\bm{s}^m$. Authors have implemented this strategy of extracting error term values and computing Shapley values for each patient in both linear non-Gaussian and heteroscedastic noise models \cite{Strobl22a,Strobl22b}. The strategy has also been generalized to confounding in the linear non-Gaussian case \cite{Strobl23}. The authors however only framed root causes in interventionalist terms. We instead strengthen the justification of the strategy by deriving the same Shapley values based on fixed factual data using a counterfactual argument.

\section{Comparison to Prior Work}

Other authors have proposed counterfactual formulations that are inappropriate for identifying patient-specific root causes of disease. \cite{Budhathoki21} for example identifies changes in the marginal distribution $\mathbb{P}(D)$, but marginalization forgoes patient-specificity. We focus on changes in the \textit{conditional} distribution $\mathbb{P}(D|\bm{E})$. \cite{Budhathoki22} quantifies the probability of encountering an event more extreme than the one observed. Their method therefore identifies root causes of having symptoms worse than a given
patient. We do not want to eliminate just the worse symptoms of a patient but \textit{all} of his symptoms, so we instead identify root causes irrespective of symptom severity.

\cite{Budhathoki22,Budhathoki21} also assume knowledge of the counterfactual distribution $\mathbb{P}(\bm{E}^*)$, but it is not clear how to choose such distributions in biomedical applications. The authors in \cite{Von23} specify the formulation of \cite{Budhathoki22} with backtracking counterfactuals but leave the backtracking conditional unspecified so that the user is still unable to identify $\mathbb{P}(\bm{E}^*)$. We on the other hand explicitly choose $\mathbb{P}(\bm{E}^*=\bm{e}| \bm{e}) = 1$ with backtracking counterfactuals in order to recover the intuitive backtracking strategy employed by clinicians on factual data.

The strategies of \cite{Budhathoki22,Budhathoki21} carry other shortcomings. First, the authors assume that the diagnosis $D$ corresponds to a deterministic function of $\bm{X}$, even though the diagnosis $D$ is stochastic in practice due to imperfect reliability. Moreover, their root causal contribution measures depend on disease prevalence, whereas ours does not. Finally, their proposed Shapley values require enumeration over all possible sets or Monte Carlo sampling, whereas we can leverage existing algorithms for fast approximation even in the nonlinear case \cite{Lundberg17,Lundberg18}. Our formulation thus allows a noisy label, accommodates changes in disease prevalence and admits fast computation in addition to exactly specifying the values of $\bm{E}^*$.

\section{Conclusion}
We climbed to the third rung of Pearl's Ladder of Causation by defining patient-specific root causes of disease using counterfactuals. We justified our approach by first defining a patient-specific root cause of $D$ as a root vertex and appreciable cause of $D$. We then exploited this definition in interventional counterfactuals by quantifying the change in likelihood of developing disease between the factual and counterfactual worlds with $\mathbb{P}(\bm{E}_{\bm{V}}^*)=\mathbb{P}(\bm{E}_{\bm{V}})$. Next, we connected interventional counterfactuals with backtracking counterfactuals using the backtracking conditional $\mathbb{P}(\bm{E}^* = \bm{e} | \bm{e}) = 1$ that preserves the error term values of the factual world. We finally quantified root causal contributions of individual variables regardless of set choice using the Shapley values of \cite{Lundberg17}. Our approach accommodates noisy labels, adapts to changes in disease prevalence, admits fast computations of the Shapley values and -- most importantly -- matches clinical intuition on fixed factual data. 

\bibliography{biblio}


\begin{thebibliography}{13}


\ifx \showCODEN    \undefined \def \showCODEN     #1{\unskip}     \fi
\ifx \showDOI      \undefined \def \showDOI       #1{#1}\fi
\ifx \showISBNx    \undefined \def \showISBNx     #1{\unskip}     \fi
\ifx \showISBNxiii \undefined \def \showISBNxiii  #1{\unskip}     \fi
\ifx \showISSN     \undefined \def \showISSN      #1{\unskip}     \fi
\ifx \showLCCN     \undefined \def \showLCCN      #1{\unskip}     \fi
\ifx \shownote     \undefined \def \shownote      #1{#1}          \fi
\ifx \showarticletitle \undefined \def \showarticletitle #1{#1}   \fi
\ifx \showURL      \undefined \def \showURL       {\relax}        \fi
\providecommand\bibfield[2]{#2}
\providecommand\bibinfo[2]{#2}
\providecommand\natexlab[1]{#1}
\providecommand\showeprint[2][]{arXiv:#2}

\bibitem[Budhathoki et~al\mbox{.}(2021)]%
        {Budhathoki21}
\bibfield{author}{\bibinfo{person}{Kailash Budhathoki},
  \bibinfo{person}{Dominik Janzing}, \bibinfo{person}{Patrick Bl{\"o}baum},
  {and} \bibinfo{person}{Hoiyi Ng}.} \bibinfo{year}{2021}\natexlab{}.
\newblock \showarticletitle{Why did the distribution change?}. In
  \bibinfo{booktitle}{\emph{International Conference on Artificial Intelligence
  and Statistics}}. PMLR, \bibinfo{pages}{1666--1674}.
\newblock


\bibitem[Budhathoki et~al\mbox{.}(2022)]%
        {Budhathoki22}
\bibfield{author}{\bibinfo{person}{Kailash Budhathoki}, \bibinfo{person}{Lenon
  Minorics}, \bibinfo{person}{Patrick Bl{\"o}baum}, {and}
  \bibinfo{person}{Dominik Janzing}.} \bibinfo{year}{2022}\natexlab{}.
\newblock \showarticletitle{Causal structure-based root cause analysis of
  outliers}. In \bibinfo{booktitle}{\emph{International Conference on Machine
  Learning}}. PMLR, \bibinfo{pages}{2357--2369}.
\newblock


\bibitem[Lauritzen et~al\mbox{.}(1990)]%
        {Lauritzen90}
\bibfield{author}{\bibinfo{person}{Steffen~L Lauritzen},
  \bibinfo{person}{A~Philip Dawid}, \bibinfo{person}{Birgitte~N Larsen}, {and}
  \bibinfo{person}{H-G Leimer}.} \bibinfo{year}{1990}\natexlab{}.
\newblock \showarticletitle{Independence properties of directed Markov fields}.
\newblock \bibinfo{journal}{\emph{Networks}} \bibinfo{volume}{20},
  \bibinfo{number}{5} (\bibinfo{year}{1990}), \bibinfo{pages}{491--505}.
\newblock


\bibitem[Lundberg et~al\mbox{.}(2018)]%
        {Lundberg18}
\bibfield{author}{\bibinfo{person}{Scott~M Lundberg},
  \bibinfo{person}{Gabriel~G Erion}, {and} \bibinfo{person}{Su-In Lee}.}
  \bibinfo{year}{2018}\natexlab{}.
\newblock \showarticletitle{Consistent individualized feature attribution for
  tree ensembles}.
\newblock \bibinfo{journal}{\emph{arXiv preprint arXiv:1802.03888}}
  (\bibinfo{year}{2018}).
\newblock


\bibitem[Lundberg and Lee(2017)]%
        {Lundberg17}
\bibfield{author}{\bibinfo{person}{Scott~M Lundberg} {and}
  \bibinfo{person}{Su-In Lee}.} \bibinfo{year}{2017}\natexlab{}.
\newblock \showarticletitle{A unified approach to interpreting model
  predictions}. In \bibinfo{booktitle}{\emph{Proceedings of the 31st
  International Conference on Neural Information Processing Systems}}.
  \bibinfo{pages}{4768--4777}.
\newblock


\bibitem[Pearl(2009)]%
        {Pearl09}
\bibfield{author}{\bibinfo{person}{Judea Pearl}.}
  \bibinfo{year}{2009}\natexlab{}.
\newblock \bibinfo{booktitle}{\emph{Causality}}.
\newblock \bibinfo{publisher}{Cambridge University Press},
  \bibinfo{address}{Cambridge}.
\newblock


\bibitem[Peters et~al\mbox{.}(2014)]%
        {Peters14}
\bibfield{author}{\bibinfo{person}{Jonas Peters}, \bibinfo{person}{Joris~M
  Mooij}, \bibinfo{person}{Dominik Janzing}, {and} \bibinfo{person}{Bernhard
  Sch{\"o}lkopf}.} \bibinfo{year}{2014}\natexlab{}.
\newblock \showarticletitle{Causal discovery with continuous additive noise
  models}.
\newblock \bibinfo{journal}{\emph{Journal of Machine Learning Research}}
  (\bibinfo{year}{2014}).
\newblock


\bibitem[Piaget(2014)]%
        {Piaget14}
\bibfield{author}{\bibinfo{person}{Jean Piaget}.}
  \bibinfo{year}{2014}\natexlab{}.
\newblock \bibinfo{booktitle}{\emph{Studies in Reflecting Abstraction}}.
\newblock \bibinfo{publisher}{Psychology Press}, \bibinfo{address}{United
  States}.
\newblock


\bibitem[Shimizu et~al\mbox{.}(2011)]%
        {Shimizu11}
\bibfield{author}{\bibinfo{person}{Shohei Shimizu}, \bibinfo{person}{Takanori
  Inazumi}, \bibinfo{person}{Yasuhiro Sogawa}, \bibinfo{person}{Aapo
  Hyv{\"a}rinen}, \bibinfo{person}{Yoshinobu Kawahara},
  \bibinfo{person}{Takashi Washio}, \bibinfo{person}{Patrik~O Hoyer}, {and}
  \bibinfo{person}{Kenneth Bollen}.} \bibinfo{year}{2011}\natexlab{}.
\newblock \showarticletitle{DirectLiNGAM: A direct method for learning a linear
  non-Gaussian structural equation model}.
\newblock \bibinfo{journal}{\emph{The Journal of Machine Learning Research}}
  \bibinfo{volume}{12} (\bibinfo{year}{2011}), \bibinfo{pages}{1225--1248}.
\newblock


\bibitem[Strobl and Lasko(2022a)]%
        {Strobl22a}
\bibfield{author}{\bibinfo{person}{Eric~V. Strobl} {and}
  \bibinfo{person}{Thomas~A. Lasko}.} \bibinfo{year}{2022}\natexlab{a}.
\newblock \showarticletitle{Identifying Patient-Specific Root Causes of
  Disease}. In \bibinfo{booktitle}{\emph{Proceedings of the 13th ACM
  International Conference on Bioinformatics, Computational Biology and Health
  Informatics}} (Northbrook, Illinois) \emph{(\bibinfo{series}{BCB '22})}.
  \bibinfo{publisher}{Association for Computing Machinery},
  \bibinfo{address}{New York, NY, USA}, Article \bibinfo{articleno}{18},
  \bibinfo{numpages}{10}~pages.
\newblock
\showISBNx{9781450393867}


\bibitem[Strobl and Lasko(2022b)]%
        {Strobl22b}
\bibfield{author}{\bibinfo{person}{Eric~V. Strobl} {and}
  \bibinfo{person}{Thomas~A. Lasko}.} \bibinfo{year}{2022}\natexlab{b}.
\newblock \showarticletitle{Identifying Patient-Specific Root Causes with the
  Heteroscedastic Noise Model}.
\newblock \bibinfo{journal}{\emph{arXiv preprint arXiv:2205.13085}}
  (\bibinfo{year}{2022}).
\newblock


\bibitem[Strobl and Lasko(2023)]%
        {Strobl23}
\bibfield{author}{\bibinfo{person}{Eric~V Strobl} {and}
  \bibinfo{person}{Thomas~A Lasko}.} \bibinfo{year}{2023}\natexlab{}.
\newblock \showarticletitle{Sample-Specific Root Causal Inference with Latent
  Variables}.
\newblock \bibinfo{journal}{\emph{Causal Learning and Reasoning}}
  (\bibinfo{year}{2023}).
\newblock


\bibitem[von K{\"u}gelgen et~al\mbox{.}(2023)]%
        {Von23}
\bibfield{author}{\bibinfo{person}{Julius von K{\"u}gelgen},
  \bibinfo{person}{Abdirisak Mohamed}, {and} \bibinfo{person}{Sander Beckers}.}
  \bibinfo{year}{2023}\natexlab{}.
\newblock \showarticletitle{Backtracking Counterfactuals}.
\newblock \bibinfo{journal}{\emph{Causal Learning and Reasoning}}
  (\bibinfo{year}{2023}).
\newblock


\end{thebibliography}
\end{document}